%% file: root.tex
\documentclass[letterpaper, 10 pt, conference]{ieeeconf}  

\IEEEoverridecommandlockouts                              

\usepackage{amsmath} 
\usepackage{bm}
\usepackage[nolist]{acronym}
\usepackage{graphicx}
\usepackage{comment}
\usepackage{cite}
\usepackage{booktabs}
\usepackage{units}

\usepackage{subfig}

\usepackage{tikz}
\newcommand{\etal}{\textit{et al}.}

\title{\LARGE \bf
Path-Constrained State Estimation for Rail Vehicles
}

\author{Cornelius von Einem$^{1,\ast}$, Andrei Cramariuc$^{2,\ast}$, Roland Siegwart$^{1}$, Cesar Cadena$^{1}$ and Florian Tschopp$^{3}$
\thanks{$^\ast$Authors contributed equally to this work}
\thanks{$^1$Authors are members of the Autonomous Systems Lab, ETH Zurich, Switzerland; {\tt\small \{firstname.lastname\}@mavt.ethz.ch}}
\thanks{$^2$Author is a member of the Robotics Systems Lab, ETH Zurich, Switzerland but the work was done while the author was a member of $^1$; {\tt\small \{firstname.lastname\}@mavt.ethz.ch}}%
\thanks{$^3$Author is with 
Arrival Ltd., London, United Kingdom; {\tt\small tschopp@arrival.com}}%
\thanks{This work was supported by the ETH Mobility Initiative under the project \textit{LROD-ADAS}.}%
}

\begin{document}

\maketitle
\thispagestyle{empty}
\pagestyle{empty}

\begin{abstract}

Globally rising demand for transportation by rail is pushing existing infrastructure to its capacity limits, necessitating the development of accurate, robust, and high-frequency positioning systems to ensure safe and efficient train operation. 
As individual sensor modalities cannot satisfy the strict requirements of robustness and safety, a combination thereof is required. 
We propose a path-constrained sensor fusion framework to integrate various modalities while leveraging the unique characteristics of the railway network.
To reflect the constrained motion of rail vehicles along their tracks, the state is modeled in 1D along the track geometry.
We further leverage the limited action space of a train by employing a novel multi-hypothesis tracking to account for multiple possible trajectories a vehicle can take through the railway network. 
We demonstrate the reliability and accuracy of our fusion framework on multiple tram datasets recorded in the city of Zurich, utilizing \acl{VIO} for local motion estimation and a standard \acs{GNSS} for global localization. 
We evaluate our results using ground truth localizations recorded with a \acs{RTK}-\acs{GNSS}, and compare our method to standard baselines. 
A \acl{RMSE} of \unit[4.78]{m} and a track selectivity score of up to \unit[94.9]{$\%$} have been achieved.

\end{abstract}

\acresetall

\section{INTRODUCTION}

Railway passenger numbers are steadily increasing with increasing environmental awareness and widespread demand for more personal mobility. 
Simultaneously, governments are pushing to increase global trade volumes via rail transport rather than truck and air freight. 
These factors push the current railway networks closer to their maximum operational capacity.
New modes of operation and technologies are needed to facilitate more trains without building new tracks.
Most railway network safety systems, such as the \ac{ETCS} Level 0-2~\cite{ETCSEngineers2011}, operate using fixed block interlocking.
In \ac{ETCS}, motion authority is granted to incoming vehicles only if entire track segments are guaranteed to be free using sparse track-side infrastructure beacons (Balises). 
A more efficient moving block strategy would permit for a denser operation of the network but would require accurate and continuous knowledge of the position of each train~\cite{beuginSimulationbasedEvaluationDependability2012, maraisSurveyGNSSBasedResearch2017, albrechtPreciseReliableTrain2013}.
To ensure the safe operation of the network, we need to know the specific track the vehicle is located on and where along the track it is. 
This is to guarantee sufficient braking distances, as well as to detect wrong turns or either missed or faulty signals at switches.

\begin{figure}
    \centering
    \includegraphics[width=0.9\linewidth]{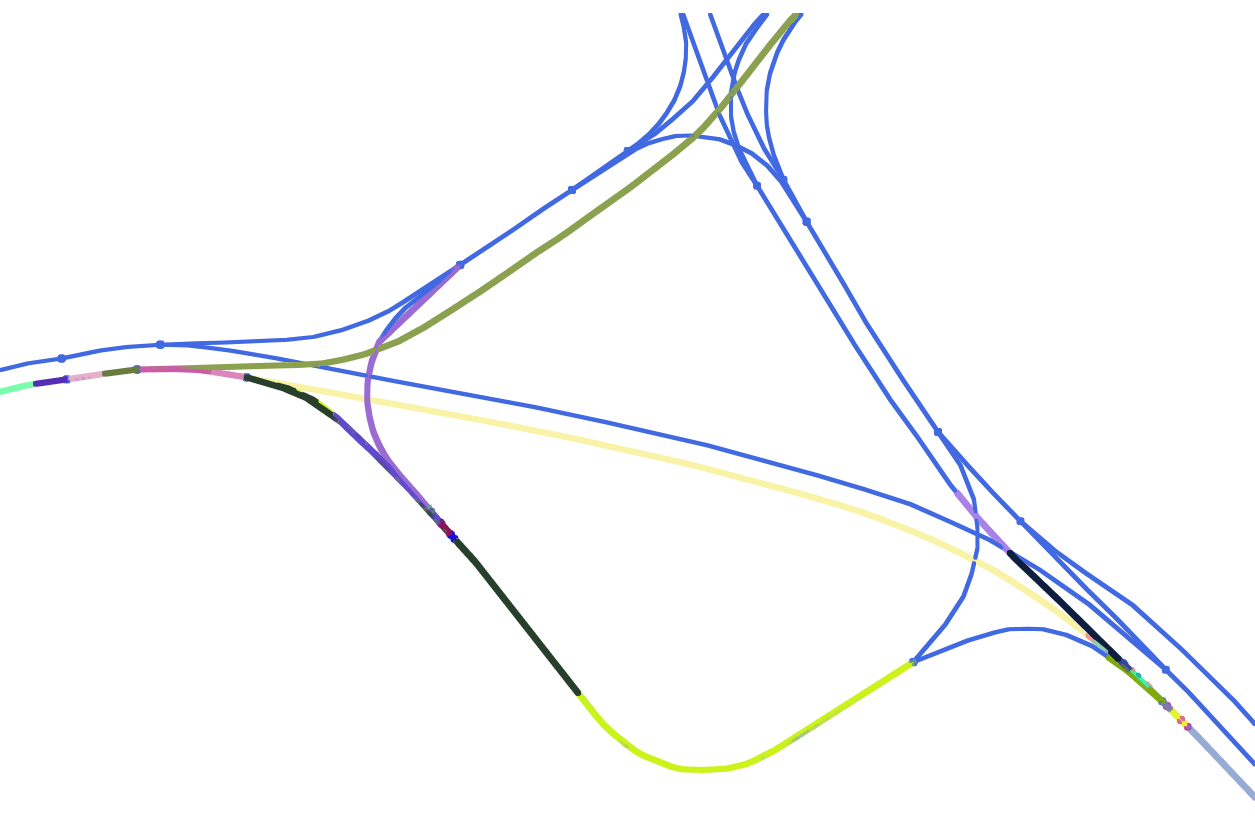}
    \caption{Multiple filter hypotheses following each track (blue) through the rail network. Each color represent one motion trajectory hypothesis.}
    \label{fig:teaser}
\end{figure}

For railways, a lot of focus has been on the use of \ac{GNSS}\cite{beuginSimulationbasedEvaluationDependability2012, amatettiFutureGenerationRailway2022} for global positioning. 
However, \ac{GNSS} has many modes of failure, and for higher levels of reliability, multiple independent modalities need to be combined~\cite{liuTestEvaluationGNSSbased2020}.
Typical sensor fusion systems employ some variant of filtering, such as a particle filter or a \ac{KF}, to integrate various modalities into a common state space representation.
However, using standard approaches that represent the state as a 3~\acp{DoF} or 6~\ac{DoF} leads to estimates that are in reality impossible, \textit{i.e.} a train can not be outside the tracks.
These approaches also do not accurately reflect the uncertainty of which track the vehicle is most likely on, just that it is somewhere off-track.
Many commonly used sensors, including \ac{GNSS}, also do not provide measurements that are constrained to the tracks.
Currently, existing sensor fusion approaches for rail networks are map-agnostic and do not leverage this extremely useful information~\cite{oteguiSurveyTrainPositioning2017, sengupta2020choice}.

In this work, we specifically address railway networks and propose a novel state estimation framework, tailored to the constraints of the environment.
We design our own \ac{EKF} to track the position of a singular vehicle over time. 
%
%
Since rail vehicles can only move in a line, we model the state space of the filter as the distance traveled along a specific track, where we define a track as the rail segment between two intersections.
In combination with known and publicly available track maps, a full 6\ac{DoF} pose of the vehicle can always be recovered. 
We propose a novel way to transfer sensor measurements from different modalities into this 1D track space to update the filter state.
In doing this projection, we also gain additional insight into the correctness of our estimate from the map information.
We implicitly check that the geometry of the estimated motion matches the shape of the track, \textit{e.g.} if the motion estimate goes in a line but on the map we have a curve, the uncertainty of our estimate should increase.

Ordinary \acp{EKF} can only track a singular state hypothesis, which does not reflect well the various possible trajectories a vehicle can take through the network.
\textit{E.g.,} we do not want an error at an intersection to cause the train to be localized on the adjacent track. 
We represent this by utilizing a collection of \acp{EKF} that track multiple possible movement hypotheses, sampling new filters at intersections and removing ones that have diverged due to disagreeing sensor updates, as shown in Figure~\ref{fig:teaser}. 
To evaluate the agreement between individual sensor measurements and each state hypothesis, we integrate a cost function and outlier rejection scheme into the \acp{EKF}.
We evaluate our framework extensively on real-world data from trams in the city of Z\"urich, using a \ac{GNSS} for global positioning and a \ac{VIO} algorithm for motion estimation.
However, our method remains generic and can be applied to other global and motion estimation sources, such as wheel odometers, \acp{IMU} or Balises.
To summarize, our contributions are as follows:
\begin{itemize}
    \item A path-constrained \ac{EKF} formulation and update scheme for modeling the movement of rail vehicles along a known rail network.
    \item An \ac{EKF}-based multi-hypothesis tracking scheme for observing various possible vehicle trajectories throughout the railway network. 
    \item Extensive evaluations on a real-world dataset, recorded on trams in Zurich, Switzerland. We do not only focus on correct track identification but also evaluate the accuracy of the position along the track.
\end{itemize}

\section{RELATED WORK}

\subsection{Sensor Fusion}
Sensor fusion is necessary to compensate for failure modes of different modalities and has seen two main-stream approaches.
First, filter-based approaches, such as particle filters~\cite{caron2007particle, gustafsson2010particle} and \acfp{KF}~\cite{julier2004unscented, chen2011kalman, bloesch2017two} with many variants and extensions.
Second, sliding window-based optimization approaches~\cite{leutenegger2015keyframe, mascaro2018gomsf} that create a local factor graph and use a non-linear optimizer to solve the unknown states.
Except for particle filters, none of the other approaches is implicitly able to handle multiple hypotheses, which has shown to be beneficial, for example in tracking~\cite{weng2022mtp} or landmark association~\cite{bernreiter2019multiple}.
To the best of our knowledge, our approach is the first to propose using a collection of \acfp{EKF} to track and rank multiple hypotheses simultaneously.
The only similar approach is the multi-hypothesis \ac{EKF} proposed by Tan~\etal~\cite{tan2022shape} that uses only a single filter with multiple internal hypotheses for soft shape estimation.

\subsection{Map-aided Localization}

Cars and trucks are limited to roads~\cite{heidenreichLaneSLAMSimultaneousPose2015, petrichAssessingMapbasedManeuver2014}, trains are limited to their tracks~\cite{maraisSurveyGNSSBasedResearch2017}, and even ships are partially limited to rivers or stick to shipping lanes in the open oceans~\cite{krishanthPredictionRetrodictionAlgorithms2014}. 
Knowledge of these paths (\textit{i.e.} maps) can aid the localization process to achieve more accurate and reliable results, especially when paths present unique geometries~\cite{wijesomaObservabilityPathConstrained2006}.
A common approach is so-called map-matching, where the current vehicle state (\textit{i.e.} pose) or its recent trajectory is matched to certain state possibilities on a map.
Quddus~\etal~\cite{quddusCurrentMapmatchingAlgorithms2007} analyzed various map-matching algorithms and their limitations. 
Point-to-point matching~\cite{bernsteinIntroductionMapMatching} is highly sensitive to the resolution of the map, leading to easy failure cases. 
Point-to-curve~\cite{leeConstrainedSLAMApproach2007, liuGNSSTrackmapCooperative2014, jiangNewTrainIntegrity2020, caoINSOdometerTrackmapaided2022} or curve-to-curve matching~\cite{chuGPSMEMSINS2013, cossaboomAugmentedKalmanFilter2012} methods allow for more robust state estimation independent of sensor modalities by comparing local estimates to curves in the map. 
To improve the robustness of these matching schemes, one can incrementally match larger segments~\cite{brakatsoulas2005map}, introduce voting schemes~\cite{yuanInteractiveVotingBasedMap2010}, or candidate graphs~\cite{louMapmatchingLowsamplingrateGPS2009}.
However, these approaches all model the vehicle state as being feasible in the entire Euclidean space and use map data only to post-correct the position estimates, which can lead to poor track selectivity, i.e. unreliable knowledge of which track the vehicle is located on.

In contrast, the vehicle position can be constrained from the beginning always to be located on a path.
This is achievable using a histogram filter~\cite{peng2020map}, though the downside is the discretization of the state space.
Other options include particle filters~\cite{kirubarajanTrackingGroundTargets1998, heirichBayesianTrainLocalization2016} with the drawback of complicated path-constrained particle re-sampling. 
Operating in curve coordinates along the pre-defined paths results in a natural description of the vehicle kinematics and allows for the usage of \acp{KF}~\cite{cossaboomAugmentedKalmanFilter2012, lauerTrainLocalizationAlgorithm2014, hasbergSimultaneousLocalizationMapping2012}. 
An inherent issue of these \acp{KF} is the branching of paths, where wrong choices in the intersection lead to unrecoverable failures in the state estimation.
Our proposed approach re-designs the \acs{KF}-based approach to seamlessly combine path-constrained state estimation with multi-hypothesis tracking into one unified framework aimed at rail networks.

\subsection{State Estimation for Trains}
Switching existing railway networks from a low-resolution infrastructure side localization system, using track-mounted Balises, to an accurate and continuous state estimation would enable train operators to deploy significantly more trains on the existing tracks while ensuring current levels of safety~\cite{williamsNextETCSLevel2016}.
\ac{GNSS} is an obvious choice for the global localization of trains~\cite{beuginSimulationbasedEvaluationDependability2012, amatettiFutureGenerationRailway2022}. 
However, \ac{GNSS} is highly dependent on a clear view of the sky and can easily be obstructed through jamming devices~\cite{liuTestEvaluationGNSSbased2020}, so it is not suitable for safety-critical applications and should be fused with other modalities.
Different other methods have been proposed for positioning rail vehicles, from RFID tags~\cite{kostrominovRFIDBasedNavigationSubway2020, zhengTrainIntegratedPositioning2016}, magnetic sensors~\cite{henselProbabilisticLandmarkBased2010, heirichStudyTrainSidePassive2017}, vibration signatures~\cite{heirich2013velocity}, vision~\cite{wohlfeilVisionBasedRail2011}, to LiDARs~\cite{naiTrainPositioningAlgorithm2017, daoustLightEndTunnel2016}.
Similarly, drifting local motion estimation can be done using a variety of modalities, such as \acp{IMU} and wheel odometers~\cite{allottaLocalizationAlgorithmRailway2015}, Doppler radars~\cite{sengupta2020choice}, or vision~\cite{tschoppExperimentalComparisonVisualAided2019, tschoppVisualPositioningRailway2021a}. 
Fusions of these methods have been map-agnostic~\cite{oteguiSurveyTrainPositioning2017, sengupta2020choice}, and they are still prone to failures and measurements that are off-track.
While we present our approach using a fusion of \ac{GNSS} positioning and \ac{VIO} motion estimation, our framework is agnostic to the types of modalities it fuses.
We provide the means to fuse any combination of the modalities presented here and, therefore, obtain a more precise on-track state estimation.

\section{Methodology}

In this section, our path-constrained state estimation framework is introduced. 
First, we explain the structure of the available track map and how we formulate the vehicle state space within this map. 
A custom path-constrained \ac{EKF} then propagates the vehicle state within this state space by integrating available sensor measurements. 
As a final step, we show how our framework can track multiple hypotheses simultaneously while keeping track of the most likely hypothesis for each time instant.

\subsection{Map structure}
\begin{figure}
    \centering
    \includegraphics[width=0.9\linewidth,trim={2cm 3cm 2cm 2cm},clip]{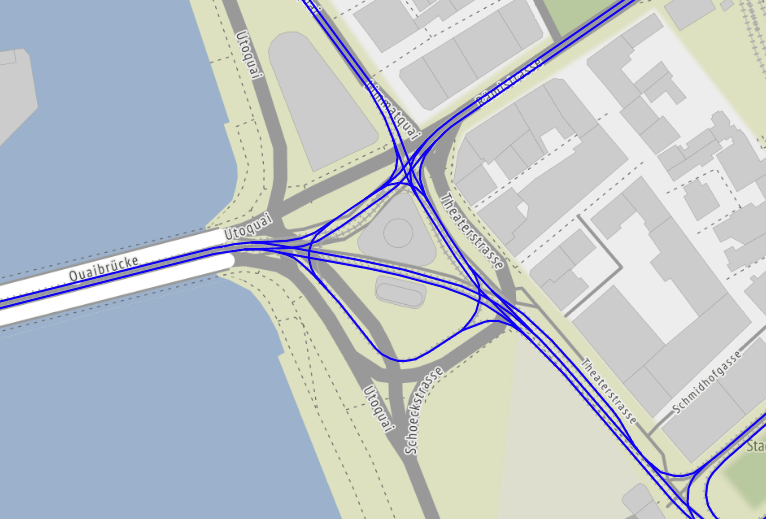}
    \caption{Tram tracks (highlighted in blue) at the Bellevue intersection in Zurich, Bellevue, showing each possible path a tram can take. Such complex intersections highlight the need for algorithms that can deal robustly with sensor noise and ambiguity.}
    \label{fig:zurich_bellevue}
\end{figure}

For railway networks, one can typically differentiate between four types of maps~\cite{winterGeneratingCompactGeometric2019a}:
\begin{itemize}
    \item \textbf{Topological track maps} describe the general connectivity within a railway network, only containing the existence of individual tracks and how these connect to one another. 
    These maps are currently used in the operation of rail networks, as the exact position of the vehicles within the network is unknown, only the track they are located on. 
    \item \textbf{Topographic track maps} in addition to the connectivity information in topological maps, also store the track course in absolute coordinates, as shown in Figure~\ref{fig:zurich_bellevue}.
    \item \textbf{Geometric track maps} are topographic maps that also contain other information about the local track geometry, such as bank angles.
    \item \textbf{Feature maps} incorporate more high-level information, such as visual landmarks or magnetic signatures, which can be utilized for the purpose of localization, maintenance, or other use cases. 
\end{itemize}
As topographic maps are easily and freely available on the internet, we use these as the basis of our state estimation system.
However, our method can be extended to more advanced map types to employ additional track information in the filter state for more accurate localization. 

Our maps, therefore, consist of small track segments which are described by a series of control points with global coordinates on a 2D plane. 
The sequence of control points defines the length, shape, and directionality of the track segment. 
The local curve-coordinates can now be defined as a continuous variable starting at the first control point and integrating along the track until the end of the segment is reached. 
At the end of the track segments are intersection points that connect to one or more additional segments, which each have their own unique ID and control points. 

\subsection{Path-constrained state space}

\begin{figure}
    \centering
    \includegraphics[width=0.9\linewidth]{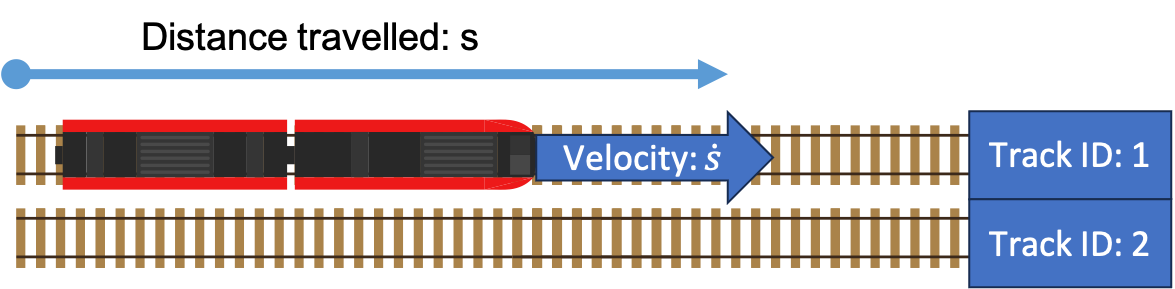}
    \caption{A track map and the state of the vehicle on the track suffice to describe the full vehicle pose in Euclidean coordinates.}
    \label{fig:state_space}
\end{figure}

To constrain the vehicle state onto the train tracks and to provide a natural description of the vehicle's kinematics, the state space is described using curve-coordinates along the local track segment. 
This means that the current state can be described as a tuple $p = \langle id,s,\dot{s} \rangle$, where $id$ is the id of the current track, $s$ is the position of the vehicle along the track in the previously described curve-coordinates, and $\dot{s}$ is the vehicle's velocity in curve-coordinates.
This parameterization can thus only describe global poses which lie on the predefined tracks, see Figure~\ref{fig:state_space}. 
The vehicle state in global Euclidean coordinates can be recovered using a function $f(\cdot)$ with the given $map$:
\begin{equation}
	\label{eq:track_to_euc_conv}
	x_w,y_w,\theta_w,\dot{x_w},\dot{y_w} = f(\text{track},s,\dot{s} \mid \text{map}),
\end{equation}
where $(x_w,y_w)$ is the vehicle position in Euclidean world coordinates, $\theta_w$ is the heading and $\dot{x_w},\dot{y_w}$ is the vehicle velocity.
The inverse can be computed as: 
\begin{equation}
    \text{track},s,\dot{s},\epsilon_{pos},\epsilon_{vel} = \rho(x_w,y_w,\dot{x_w},\dot{y_w} \mid \text{map}),
\end{equation}
where $\epsilon_{pos}$ and $\epsilon_{vel}$ are the error terms describing the distance or the misalignment to the track, respectively.

\subsection{Sensor Fusion}

The sensor fusion part is implemented as an \ac{EKF} within the space of one train track.
Assuming a constant velocity model, the state space on one track is represented as: 
\begin{equation}
    \bm{x}=\begin{bmatrix}
s\\
\dot{s}
\end{bmatrix}
\end{equation}
where $s$ and $\dot{s}$ are the traveled distance and velocity along that track, respectively. 
As the track itself is also an important part of the vehicle's state, it needs to be included as well, which is described in Section~\ref{sec:multi-hypothesis} and omitted here for simplicity. 

\subsubsection{Filter propagation}
\label{sec:process_model}
The standard state transition and observation models of an \ac{EKF} are:
\begin{align}
\begin{split}
    \bm{x}_k &= f(\bm{x}_{k-1},\bm{u}_k)+\bm{Q}_k \\
    \bm{z}_k &= h(\bm{x}_k) + \bm{R}_k,
\end{split}
\end{align}
where $\bm{x}_k$ represents the current state, $\bm{u}_k$ the control input, $\bm{z}_k$ the measurement, $\bm{Q}_k$ the process noise, and $\bm{R}_k$ the measurement noise.
The general motion of the rail vehicle in 6\ac{DoF} is non-linear due to the forces applied by the train tracks onto the train, restricting its freedom of motion. 
The linearization of this process model happens indirectly through the choice of curve-coordinates and allows for a simple formulation of the state transition in 1D coordinates. 
As we do not consider a control input at the moment, this simplifies the process model to:
\begin{align}
\begin{split}
    \bm{x}_{k|k-1} &= \bm{F} \hat{\bm{x}}_{k-1|k-1}\\
    \bm{P}_{k|k-1} &= \bm{F}_k \bm{P}_{k-1|k-1} \bm{F}_k^T + \bm{Q}_k \\
     \bm{F} &= \begin{bmatrix}
1 & \delta t\\
0 & 1
\end{bmatrix}, \label{eq:process_model}
\end{split}
\end{align}
where $\bm{F}$ is the transition model, $\bm{P}$ the state covariance and $\bm{Q}$ the process noise.

\subsubsection{Measurement update}
Due to the special state space formulation, measurement updates will be non-linear as they are typically reported in Euclidean coordinates and need to be transformed into curve-coordinates. 
\ac{GNSS} position updates, for example, will most likely lie somewhere in the vicinity of the tracks, but never precisely on a 1D track. 
This update can be integrated as the conditional probability distribution of the measurement given that the state has to be located on the tracks. 
Assuming a straight track and a symmetric Gaussian distribution, this conditional distribution is another normal distribution centered around the mean point projected onto the track, with the same standard deviation. 
When projecting a measurement onto a track, its coordinate system can, therefore, be changed from Euclidean to curve-coordinates. 

\begin{figure}
    \centering
\resizebox{\linewidth}{!}{%
\begin{tikzpicture}[]
    \draw[->, thick] (0,-1.0) -- (0,2.2) node (yaxis) [left] {\text{Cost}};
    \draw[->, thick] (-0.1,0) -- (7.0,0) node (xaxis) [below] {$\epsilon$};
    \draw[thick] (0.1,-0.8) -- (-0.1,-0.8) node (khp) [left] {$-KHP$};
    \draw[thick] (0.1,0.5) -- (-0.1,0.5) node (khp) [left] {$\sigma_{min}$};
    \draw[thick] (0.1,1.5) -- (-0.1,1.5) node (khp) [left] {$\sigma_{max}$};

    \draw[red, line width=0.75mm] (0,-0.8) -- (2,-0.8);
    \draw[red, line width=0.75mm] (2,-0.8) -- (2,0.5);
    \draw[red, line width=0.75mm] (2,0.5) -- (5,1.5);
    \draw[red, line width=0.75mm] (5,1.5) -- (6.5,1.5);
    \draw[thick] (2.0,-0.1) -- (2.0,0.1) node (kS) [below right] {$\kappa\sqrt{S}$};
    \draw[thick] (5.0,-0.1) -- (5.0,0.1) node (kS2) [below] {$\alpha\kappa\sqrt{S}$};

\end{tikzpicture}
}

    \caption{Measurement acceptance gate cost function for outlier rejection.}
    \label{fig:acceptance_gate}

\end{figure}
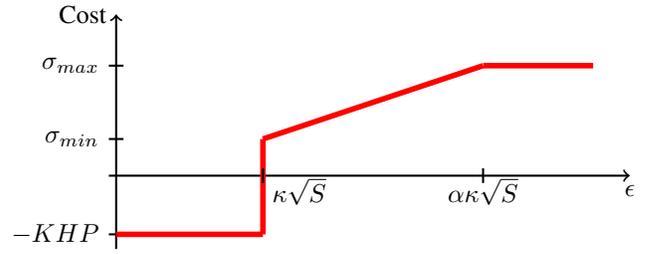

The second key part of our measurement update step is an outlier rejection scheme to remove noisy measurements. 
This is implemented in the form of a measurement acceptance gate~\cite{acceptancegate2014}. 
The projection of a measurement from Euclidean space to curve coordinates, similar to the projection of one vector onto another, also results in an error term $\epsilon$, that is perpendicular to the tracks. 
Based on the relation of this measurement term and the innovation covariance $\bm{S}$, a measurement is either accepted or rejected, and the state covariance is accordingly either updated or punished. 
If the error term $\epsilon$ is smaller than $\kappa\sqrt{\bm{S}}$, where $\kappa$ is a tuning parameter, the measurement is accepted. 
If the measurement is rejected, no measurement update is performed, and instead a punishment is applied to the state covariance. 
This punishment increases linearly from $\sigma_{min}$ to $\sigma_{max}$ until a second threshold of $\alpha\kappa\sqrt{S}$ is reached, after which only the maximum punishment is added to the state covariance. 
This can be expressed mathematically as shown in Equation~\ref{eq:acceptance_gate} and is visualized in Figure~\ref{fig:acceptance_gate}.

The resulting measurement update equations are thus: 
\begin{align}
\begin{split}
    \bm{y}_k &= \rho(\bm{z}_k) - \bm{x}_{k|k-1} \\
    \bm{S}_k &= H_k P_{k|k-1} H_k^T + R_k \\
    \bm{K}_k &= P_{k|k-1}H_k^T S_k^{-1} \\
    \bm{x}_{k|k} &= \bm{x}_{k|k-1} + \bm{K}_k \bm{y}_k    
\end{split} \\
\begin{split}
    \bm{m} &=  \frac{\sigma_{max} - \sigma_{min}}{(\alpha-1)\kappa\sqrt{\bm{S}}} \\
    \bm{P}_{k|k} &= \begin{cases}
  (\mathbf{I}   - K_k H_k)P_{k|k-1}  & \epsilon \leq \kappa \sqrt{S} \\
  P_{k|k-1} +\bm{m}(x-\kappa\sqrt{\bm{S}})+\sigma_{min}  & \text{else} \\
  P_{k|k-1} + \sigma_{max} & \epsilon > \alpha \kappa \sqrt{S} \label{eq:acceptance_gate}
\end{cases}
\end{split}
\end{align}

where $\bm{H}$ is the observation model, selecting the parts of the state space which are observable by a specific sensor. 
For an odometry update, this is $\bm{H}=\begin{bmatrix}
0 & 1
\end{bmatrix}$
and for \ac{GNSS} the reverse.

Using this, \ac{GNSS} updates can be integrated easily.
The projection function $\rho(\cdot)$ finds the closest point to the current track, which is then used for the sensor update. 
This distance to the track is the error term $\epsilon$.

The integration of odometry requires a few extra steps, since we utilize drifting \ac{VIO}.
The projection function $\rho(\cdot)$ computes the velocity component parallel to the track, at the current position estimate, while the perpendicular component represents $\epsilon$, and is used for the outlier rejection step. 
As the motion estimate will drift over time, it will be increasingly misaligned from the true motion of the vehicle, leading ultimately to a failure of the \ac{EKF}.
However, based on the current track geometry around the current position estimate, we can realign the odometry to correct this drift. 
This is done in periodic intervals of \unit[100]{m} to reduce the impact of drift, but not in close vicinity to switches, as there are multiple possible position hypothesis with severely different vehicle orientations.
This will be elaborated in further detail in Section~\ref{sec:transitioning}.

\subsection{Multi-Hypothesis tracking}
\label{sec:multi-hypothesis}

The previously described path-constrained \ac{EKF} is capable of estimating the vehicle's state along a singular track. 
For a real railway network, this is not sufficient as there are often multiple parallel tracks, switches, and crossings.
There, even small amounts of sensor noise can lead to ambiguity, especially between parallel tracks.
Thus, we propose to use a collection of \acp{EKF} to keep track of multiple possible hypotheses throughout the rail network.

\subsubsection{Initialization}
The first step is the initialization of the filters. 
After receiving a first global localization measurement (e.g., from \ac{GNSS}), \acp{EKF} are spawned on all tracks within a radius of \unit[100]{m}.

\subsubsection{Transition}
\label{sec:transitioning}
At each time step, all filters propagate according to their process model described in equation~\ref{eq:process_model}.
If a measurement update is available, it is applied to all filters.
Still, due to the outlier rejection and local track geometry, the resulting updates are different for each filter, leading to all filters propagating independently with differing covariances. 
Suppose a filter is near a map point that connects the current track to another one. 
In that case, an additional filter is created on the new track with equal velocities and covariance to maintain continuity in the transition. 
An exemplification of this can be seen in Figure~\ref{fig:teaser}.
The \ac{VIO} plays a significant role at these switches, as it will be more in accordance with one of the hypotheses over the other, which is why no odometry re-alignment occurs close to a switch. 

\subsubsection{Filter removal}
Filters can be removed from the collection of active filters under several circumstances: 
\begin{itemize}
    \item A filter reaches the end of its associated track. As this is also the end of the respective coordinate space, the filter cannot continue and gets deleted. 
    \item Measurements will not agree with the hypotheses maintained by some of the filters and will cause these filters to diverge. As the covariance of a filter increases above a certain threshold, the filter is retired and removed from the active collection. 
    \item The third option for removing filters is pruning. Railway tracks naturally split and merge throughout the network. This can lead to multiple hypotheses being located on the same track and close to each other. To prevent these filters from converging to one state, the filter with lower covariance is removed when two filters approach one another.
\end{itemize}

\subsubsection{\ac{MLE} selection}
At each time step, a \ac{MLE} is selected. 
This corresponds to the filter with the lowest positioning covariance at each time step. 
In addition, a low-pass filter is applied to this selection step, meaning that a certain filter needs to have the highest covariance for multiple time steps, to become the \ac{MLE}. 
The track ID and position along that track for the current \ac{MLE} is reported as the current state estimate of the entire system.

\section{Experimental Evaluation}

To evaluate the accuracy and reliability of our path-constrained state estimation system, we performed multiple experiments on real-world railway datasets. 
Furthermore, we directly compare the performance of our path-constrained estimator to a standard unconstrained \ac{EKF}.
Although a comparison to several of the other named railway state estimation frameworks would be informative, there are often specific to the utilized sensors which were not available to us.

\subsection{Datasets}
\begin{figure}
    \centering
    \includegraphics[width=0.9\linewidth,trim={0cm 0cm 5cm 0cm},clip]{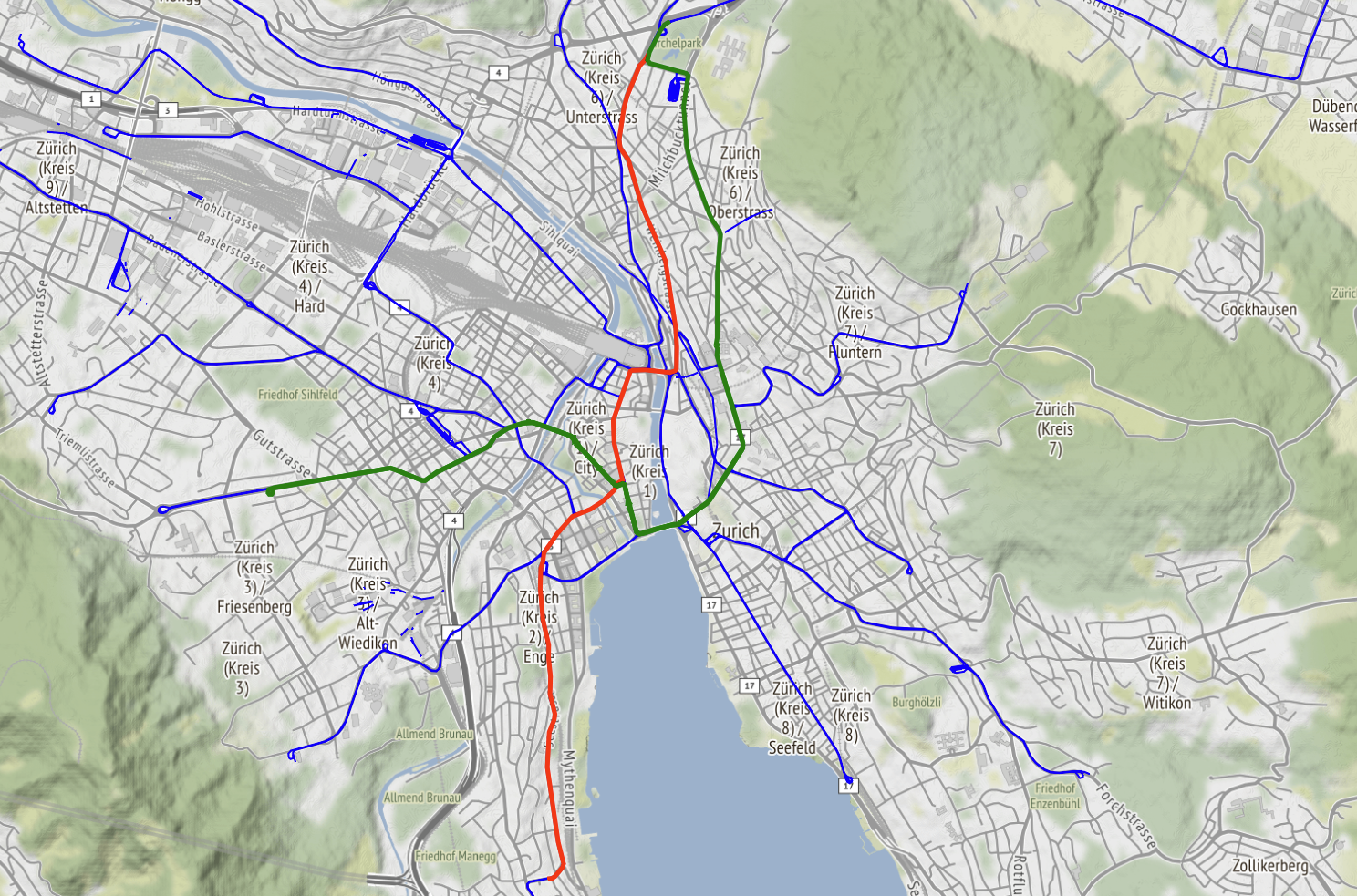}
    \caption{Map of the city of Zurich. Tram tracks are highlighted in blue, Tram 7 (\texttt{Trajectory~1}) in red and Tram 9 (\texttt{Trajectory~2} and \texttt{Trajectory~3}) in green.}
    \label{fig:dataset_trajectories}
\end{figure}

The framework was evaluated on three available datasets from trams in Zurich, Switzerland. 
\texttt{Trajectory~1} was recorded on Tram line 7, has a length of \unit[6812]{m} and a duration of \unit[26]{min}. 
\texttt{Trajectory~2} was recorded on Tram line 9, has a length of \unit[5980]{m} and a duration of \unit[23]{min}. 
\texttt{Trajectory~3} was also recorded on Tram line 9 but is significantly longer with a length of \unit[9544]{m} and a duration of \unit[45]{min}. 
All three trajectories feature a variety of different environments, ranging from the dense city center with limited \ac{GNSS} availability to outside residential areas with a clearer view of the sky.
With many intersections and high buildings obstructing the \ac{GNSS} signal, this environment is more challenging than most common railway environments.
OpenStreetMap data was used as a ground truth track map and was filtered to only contain tracks belonging to the tram network of the city of Zurich, see Figure~\ref{fig:dataset_trajectories}.

\subsection{Hardware Setup}

\subsubsection{Devices}
Localization data for the purpose of state estimation was obtained using a uBlox EVK-F9P \ac{GNSS} evaluation kit, capable of receiving the $L_1$ and $L_2$ bands of the GPS, Galileo, BeiDou, and GLONASS satellite constellations.
For \texttt{Trajectory~1}, the receiver was mounted outside the tram for a clear view of the sky. 
For \texttt{Trajectory~2} and \texttt{3}, the receiver was attached on the inside of the tram behind the glass of the window, resulting in a reduced \ac{GNSS} signal. 
Vehicle odometry was recorded using a Skybotix VI-Sensor~\cite{visensor2014}, consisting of two grayscale cameras at a resolution of $752\times480\text{px}$ and at a frame rate of \unit[25]{hz}, as well as an ADIS16488 \ac{IMU} recording at a rate of \unit[200]{hz}. 
The device was mounted to record out of a side window of the tram.
The odometry itself was computed using OKVIS~\cite{leutenegger2015keyframe} without any post-processing and thus significantly drifts over time. 
Ground truth positioning data was obtained using a PIKSI Multi \ac{RTK}-\ac{GNSS}, whose antenna was mounted on the outside of the tram for optimal positioning accuracy. 
As a post-processing step, the ground truth positioning data was projected on the ground truth track in the map, which is known from the public transportation map. 
Even though the state estimation system is significantly more complex than comparable baselines, it runs in real-time on a single core of an Intel Core i7-10875H CPU, not including the \ac{VIO} as this is not a direct part of the framework,

\subsubsection{Calibration}
The intrinsic camera parameters of the VI-Sensor have been calibrated using the \textit{Kalibr} toolbox~\cite{furgale2013unified} and a standard calibration board. 
At the time of recording, no appropriate calibration procedures were available for calibrating the VI-sensor and the various \ac{GNSS} antennas relative to the train. 
Relative orientations were therefore obtained through manual tuning, while relative translations have been neglected and were accepted as a minimal source of error in the final evaluation result.
These errors apply to all methods equally, and therefore the comparison remains fair.

\subsection{Baselines}
Due to the limited public availability of other railway sensor fusion frameworks, we compare our pipeline against two simple baselines:
\begin{itemize}
    \item \textit{RAW-\ac{GNSS}:} a first baseline is the directly available unprocessed \ac{GNSS} measurements. Due to the low update rate of \ac{GNSS}, this does not provide the same continuity as our framework. Additionally, it provides no knowledge of the current track.
    \item \textit{Projected-\ac{GNSS}:} \ac{GNSS} measurements simply projected onto the closest track. 
    \item \textit{\ac{EKF}:} the third baseline is a standard \ac{EKF}, in order to fuse \ac{GNSS} and odometry measurements, without any prior knowledge of the train tracks. This is implemented using the \texttt{robot\_pose\_ekf} \ac{ROS} package.
    \item \textit{Projected-\ac{EKF}:} the last baseline is a standard \ac{EKF}, whose estimate is projected onto the closest track in a post-processing step. 
\end{itemize}

\subsection{Metrics}
We report two distinct metrics for each baseline and our method. 
The first standard metric is the \ac{RMSE} to evaluate the positioning accuracy. 
For positioning methods that work in non-euclidean coordinates, these are first converted to normal Euclidean coordinates, as to compute the \ac{RMSE} in the same space for all methods. 

The second metric is a track selectivity score. 
For rail vehicles, it is significantly more important to know on which track the vehicle is located, than where precisely on the track it is. 
We therefore define the track selectivity score as the percentage of time, during which the current position estimate was located on the ground truth track segment. 
As not all baselines are capable of reporting the current track, this number is only reported for some methods.

\section{Results}

\begin{figure}
    \centering
    \includegraphics[trim={0 0 0 7cm},clip, width=0.8\linewidth]{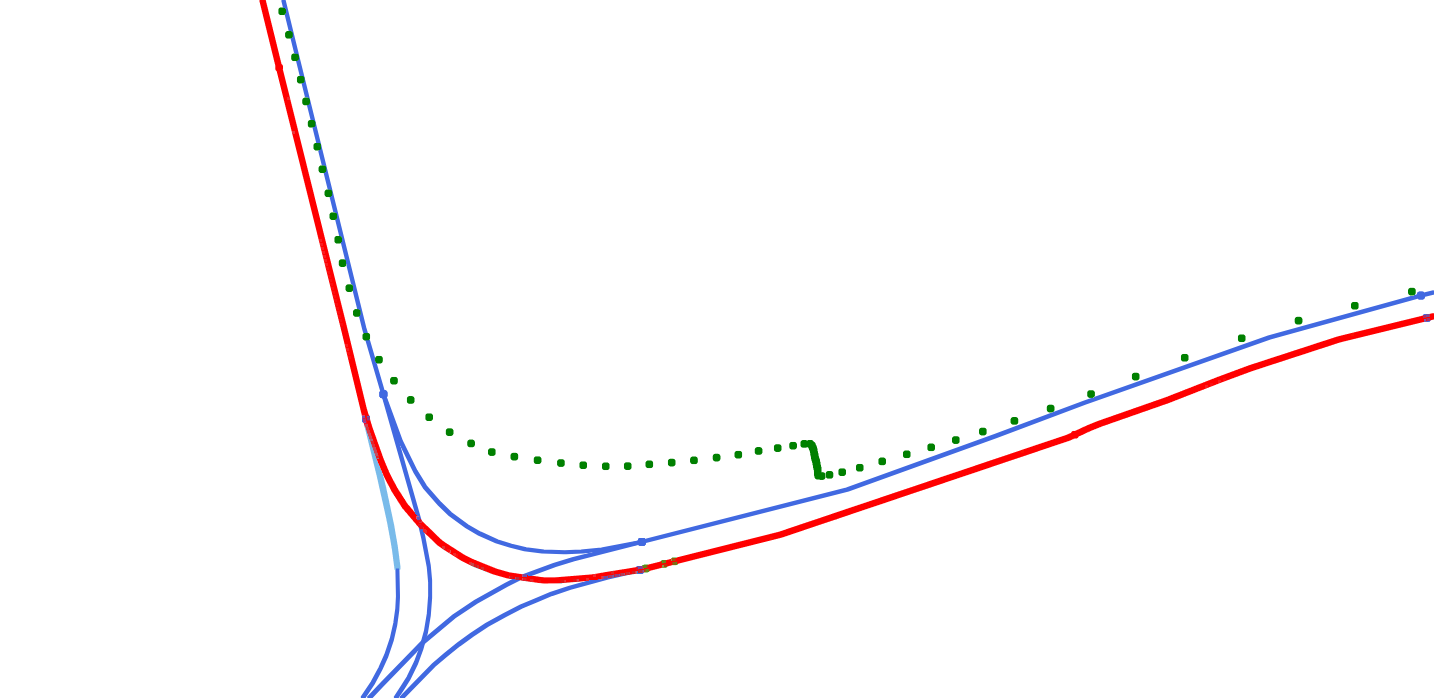}
    \caption{\ac{GNSS} signal (green) drifting severely off the track (blue), while the state estimate (red) remains on the correct track. }
    \label{fig:gnss_robustness}
\end{figure}

\begin{table*}[h!]
\centering
\caption{Positioning accuracy reported in terms of \ac{RMSE} and Track Selectivity for all methods on 3 different trajectories.}
\begin{tabular}{lcccccc}
\toprule
        & \multicolumn{2}{c}{\texttt{Trajectory~1}} & \multicolumn{2}{c}{\texttt{Trajectory~2}}    & \multicolumn{2}{c}{\texttt{Trajectory~3}}  \\ 
        & \acs{RMSE} & Track Selectivity & \acs{RMSE} & Track Selectivity & \acs{RMSE} & Track Selectivity \\ \hline
Raw-\ac{GNSS} & \unit[4.10]{m} & n/a & \unit[18.52]{m} & n/a& \unit[43.71]{m} & n/a\\
Projected-\ac{GNSS} & \textbf{\unit[2.31]{m}} & \unit[92]{$\%$} & \unit[17.73]{m} & \unit[59.7]{$\%$} & \unit[41.83]{m} & \unit[49.8]{$\%$} \\
\ac{EKF} & \unit[10.63]{m} & n/a & \unit[22.20]{m} & n/a& \unit[43.07]{m} & n/a\\
Projected-\ac{EKF} & \unit[6.27]{m} & \unit[90.6]{$\%$} & \unit[20.79]{m} & \unit[69.7]{$\%$} & \unit[41.31]{m} & \unit[50.04]{$\%$} \\ \hline
Ours & \unit[4.78]{m} & \textbf{\unit[94.9]{$\%$}} & \textbf{\unit[18.09]{m}} & \textbf{\unit[85.2]{$\%$}} & \textbf{\unit[22.63]{m}} & \textbf{\unit[81.4]{$\%$}} \\
\bottomrule
\end{tabular}
\label{table:results}
\end{table*}

\begin{figure*}%
    \centering
    \subfloat[\centering Positions of various filters (yellow, brown, violet and green) on different tracks (blue). ]{{\includegraphics[width=0.45\linewidth]{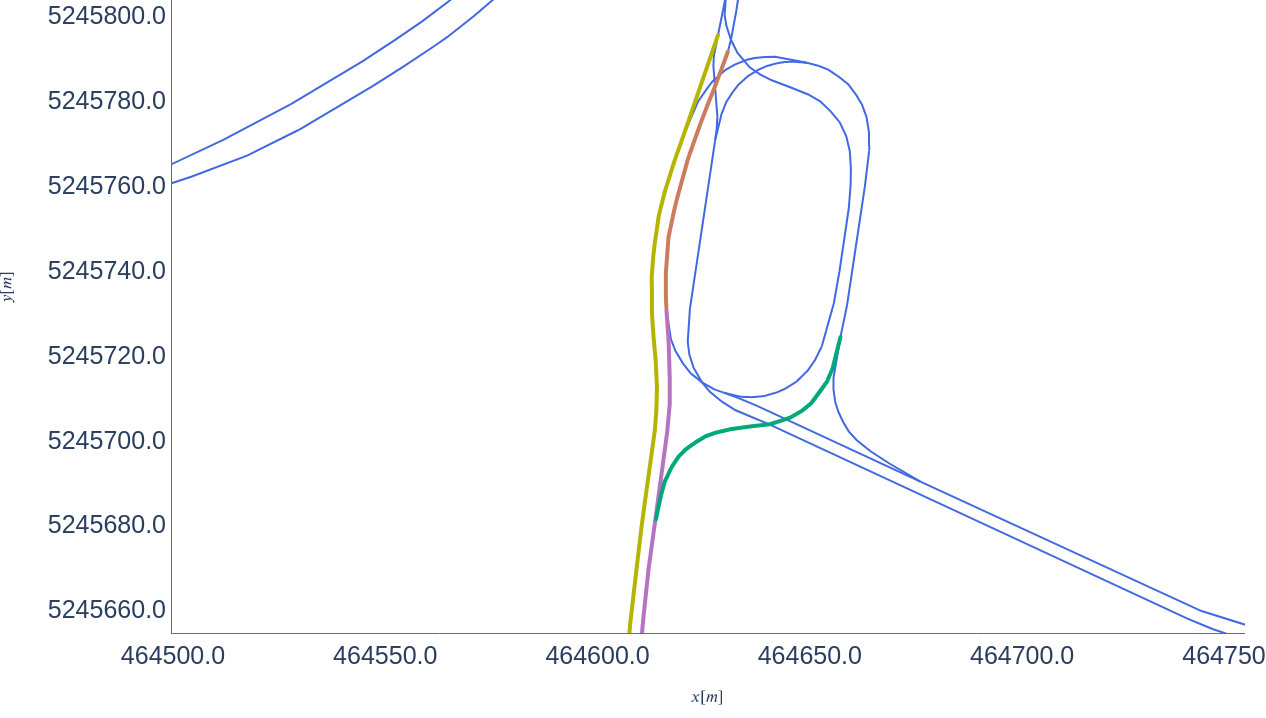} }}%
    \qquad
    \subfloat[\centering Positioning covariance of various filters over time]{{\includegraphics[width=0.45\linewidth]{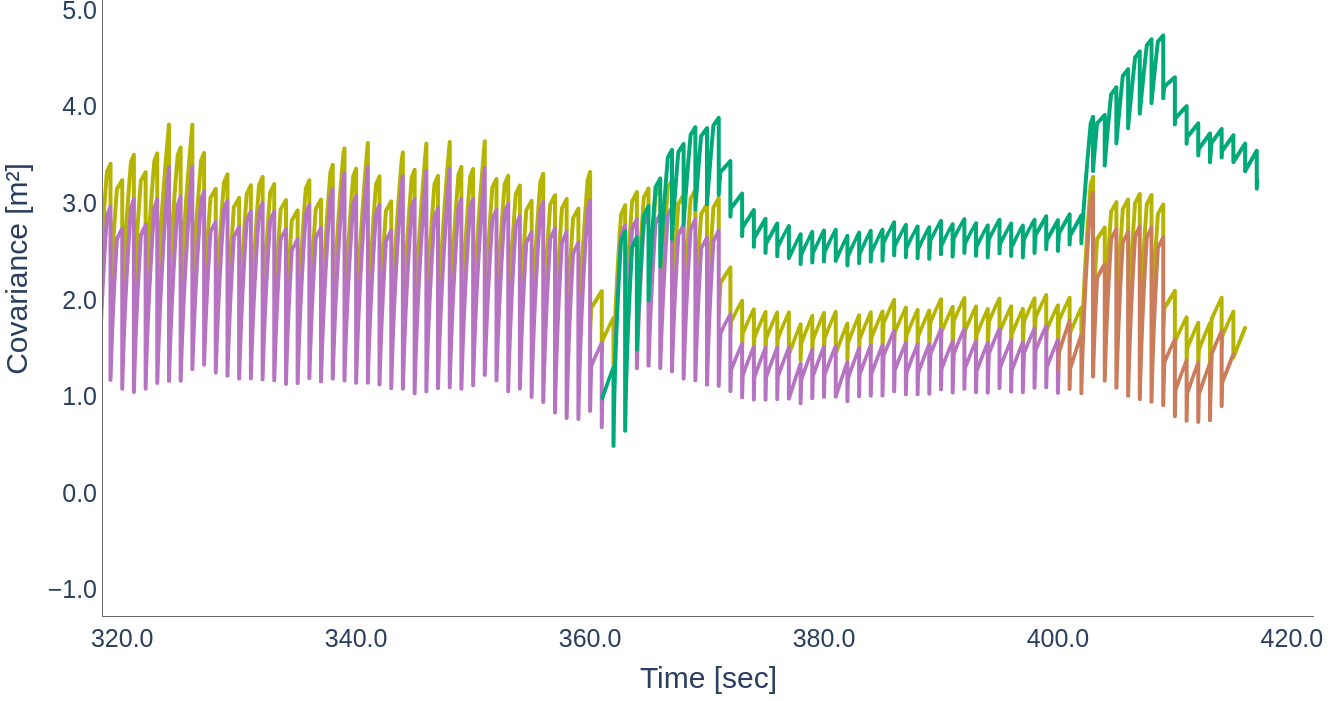} }}%
    \caption{The position of various filters can be seen on the left and there respective covariances on the right. A difference in covariance between parallel tracks is observable, as well as an increasing covariance for }%
    \label{fig:covariance_dev}%
\end{figure*}

The results for our method and our four baselines are reported in Table~\ref{table:results}.
While \ac{GNSS} can provide good localization under optimal conditions, and even a good track selectivity, the comparison between different datasets shows the high dependability on good reception. 
The \ac{VIO} on its own worked well within the Tram environment, thanks to good lighting and diverse features in urban environments, as well as limited vehicle speeds. 
It is not expected that \ac{VIO} generalizes as well to high-speed trains, but there our framework would enable the incorporation of other odometry estimation sensors.
Fusing \ac{GNSS} with \ac{VIO} in a standard \ac{EKF} results in a continuous state estimate, but does not improve the overall localization accuracy significantly, as no map data or further global localization is utilized. 
While integrating the sensor measurements in our path-constrained state estimation pipeline does not result in significant accuracy improvements over raw \ac{GNSS}, it does significantly improve the track selectivity score. 
For trains, this is the more relevant positioning metric, as it is vital to know on which track a vehicle is located, while the position along the track is only secondary. 
This can be seen in Figure~\ref{fig:gnss_robustness}, where \ac{GNSS} measurements experience significant disturbances and would even indicate the wrong track, but our fusion pipeline is robust enough to maintain the correct track estimate. 
The \ac{RMSE} is only improved by our pipeline in \texttt{Trajectory~2} and \texttt{Trajectory~3}, where the significant disturbances of the more noisy \ac{GNSS} are well compensated for using the path-constrained state-estimation.

Figure~\ref{fig:covariance_dev} highlights the changes in the covariance of various hypotheses over time.
The outlier rejection results in the divergence of some filters, such as ones taking a wrong turn at an intersection, which will then be removed from the collection of active filters. 
Furthermore, it shows that two hypothesis on parallel tracks will have a similar covariance and covariance propagation, but still with a significant gap to determine the correct track.
Bad track selectivity is predominantly observed for short distances directly after complicated intersections, where neither of the noisy sensors is providing a consistent motion estimate.

    

\section{Conclusion}

In this paper, we introduced a path-constrained state estimation framework for fusing arbitrary sensor modalities together with public map data to robustly and accurately localize rail vehicles. 
For this, we developed a novel multi-hypothesis \ac{EKF}, utilizing a curve coordinate formulation to predict the vehicle's motion along known train tracks and fuse sensor measurements into the current state. 

We demonstrate the accuracy and reliability of our pipeline using three different datasets recorded on local trams. 
Many intersections and limited \ac{GNSS} availability make this a challenging testing ground. 
Our experiments show that our pipeline significantly improves localization accuracy and robustness over common existing solutions. 

We plan to expand this work together with our event-based railway mapping framework~\cite{tschoppHoughMapIterative2021} into a full on-board high-speed railway \ac{SLAM} framework. 
Event-based vision could provide both the required odometry estimation, as well as global localization through the detection and mapping of infrastructure elements.

\addtolength{\textheight}{0cm}   






\input{acronyms}

\bibliographystyle{ieeetran}
\bibliography{root.bib}

\end{document}

%% file: acronyms.tex
\begin{acronym}
    \acro{GNSS}{Global Navigation Satellite System}
    \acro{VIO}{Visual-Inertial Odometry}
    \acro{RTK}{Real-Time Kinematic}
    \acro{IMU}{Inertial Measurement Unit}
    \acro{MAV}{Micro Aerial Vehicle}
    \acro{KF}{Kalman Filter}
    \acro{EKF}{Extended Kalman Filter}
    \acro{DoF}{Degree of Freedom}
    \acrodefplural{DoF}[DoF]{Degrees of Freedom}
    \acro{RFID}{Radio-Frequency Identification}
    \acro{RMSE}{Root Mean Square Error}
    \acro{ETCS}{European Train Control System}
    \acro{IMM}{Interacting Multiple Model}
    \acro{Lidar}{Light detection and ranging}
    \acro{SLAM}{Simultaneous Localization and Mapping}
    \acro{MLE}{Maximum Likelihood Estimate}
    \acro{ROS}{Robot Operating System}
\end{acronym}

%% file: root.bbl
\begin{thebibliography}{10}
\providecommand{\url}[1]{#1}
\csname url@rmstyle\endcsname
\providecommand{\newblock}{\relax}
\providecommand{\bibinfo}[2]{#2}
\providecommand\BIBentrySTDinterwordspacing{\spaceskip=0pt\relax}
\providecommand\BIBentryALTinterwordstretchfactor{4}
\providecommand\BIBentryALTinterwordspacing{\spaceskip=\fontdimen2\font plus
\BIBentryALTinterwordstretchfactor\fontdimen3\font minus
  \fontdimen4\font\relax}
\providecommand\BIBforeignlanguage[2]{{%
\expandafter\ifx\csname l@#1\endcsname\relax
\typeout{** WARNING: IEEEtran.bst: No hyphenation pattern has been}%
\typeout{** loaded for the language `#1'. Using the pattern for}%
\typeout{** the default language instead.}%
\else
\language=\csname l@#1\endcsname
\fi
#2}}

\bibitem{ETCSEngineers2011}
P.~Stanley, \emph{{ETCS for Engineers}}, 1st~ed.\hskip 1em plus 0.5em minus
  0.4em\relax {TZ - Verl. \& Print Gmbh}, 2011.

\bibitem{beuginSimulationbasedEvaluationDependability2012}
J.~Beugin and J.~Marais, ``{Simulation-Based Evaluation of Dependability and
  Safety Properties of Satellite Technologies for Railway Localization},''
  \emph{Transportation Research Part C: Emerging Technologies}, vol.~22, pp.
  42--57, 2012.

\bibitem{maraisSurveyGNSSBasedResearch2017}
J.~Marais, J.~Beugin, and M.~Berbineau, ``{A Survey of GNSS-Based Research and
  Developments for the European Railway Signaling},'' \emph{IEEE Transactions
  on Intelligent Transportation Systems}, vol.~18, no.~10, pp. 2602--2618,
  2017.

\bibitem{albrechtPreciseReliableTrain2013}
T.~Albrecht, K.~Luddecke, and J.~Zimmermann, ``{A Precise and Reliable Train
  Positioning System and Its Use for Automation of Train Operation},'' in
  \emph{IEEE International Conference on Intelligent Rail Transportation
  Proceedings (ICIRT)}, 2013, pp. 134--139.

\bibitem{amatettiFutureGenerationRailway2022}
C.~Amatetti, T.~Polonelli, E.~Masina, C.~Moatti, D.~Mikhaylov, D.~Amato,
  A.~Vanelli-Coralli, M.~Magno, and L.~Benini, ``{Towards the Future Generation
  of Railway Localization and Signaling Exploiting Sub-Meter RTK GNSS},'' in
  \emph{IEEE Sensors Applications Symposium (SAS)}, 2022, pp. 1--6.

\bibitem{liuTestEvaluationGNSSbased2020}
J.~Liu, J.-c. Li, B.-g. Cai, and J.~Wang, ``{Test and Evaluation of GNSS-based
  Railway Train Positioning under Jamming Conditions},'' in \emph{IEEE
  International Conference on Systems, Man, and Cybernetics (SMC)}, 2020, pp.
  1459--1464.

\bibitem{oteguiSurveyTrainPositioning2017}
J.~Otegui, A.~Bahillo, I.~Lopetegi, and L.~E. Diez, ``{A Survey of Train
  Positioning Solutions},'' \emph{IEEE Sensors Journal}, vol.~17, no.~20, pp.
  6788--6797, 2017.

\bibitem{sengupta2020choice}
M.~Sengupta, ``Choice of sensor fusion framework for train positioning
  system,'' \emph{Computers in Railways XVII: Railway Engineering Design and
  Operation}, vol. 199, pp. 53--64, 2020.

\bibitem{caron2007particle}
F.~Caron, M.~Davy, E.~Duflos, and P.~Vanheeghe, ``{Particle Filtering for
  Multisensor Data Fusion With Switching Observation Models: Application to
  Land Vehicle Positioning},'' \emph{IEEE Transactions on Signal Processing},
  vol.~55, no.~6, pp. 2703--2719, 2007.

\bibitem{gustafsson2010particle}
F.~Gustafsson, ``{Particle Filter Theory and Practice with Positioning
  Applications},'' \emph{IEEE Aerospace and Electronic Systems Magazine},
  vol.~25, no.~7, pp. 53--82, 2010.

\bibitem{julier2004unscented}
S.~J. Julier and J.~K. Uhlmann, ``{Unscented Filtering and Nonlinear
  Estimation},'' \emph{Proceedings of the IEEE}, vol.~92, no.~3, pp. 401--422,
  2004.

\bibitem{chen2011kalman}
S.~Chen, ``{Kalman Filter for Robot Vision: A Survey},'' \emph{IEEE
  Transactions on Industrial Electronics}, vol.~59, no.~11, pp. 4409--4420,
  2011.

\bibitem{bloesch2017two}
M.~Bloesch, M.~Burri, H.~Sommer, R.~Siegwart, and M.~Hutter, ``{The Two-State
  Implicit Filter Recursive Estimation for Mobile Robots},'' \emph{IEEE
  Robotics and Automation Letters (RA-L)}, vol.~3, no.~1, pp. 573--580, 2017.

\bibitem{leutenegger2015keyframe}
S.~Leutenegger, S.~Lynen, M.~Bosse, R.~Siegwart, and P.~Furgale,
  ``{Keyframe-Based Visual-Inertial SLAM Using Nonlinear Optimization},''
  \emph{The International Journal of Robotics Research (IJRR)}, vol.~34, no.~3,
  pp. 314--334, 2015.

\bibitem{mascaro2018gomsf}
R.~Mascaro, L.~Teixeira, T.~Hinzmann, R.~Siegwart, and M.~Chli, ``{GOMSF:
  Graph-Optimization Based Multi-Sensor Fusion for robust UAV Pose
  estimation},'' in \emph{2018 IEEE International Conference on Robotics and
  Automation (ICRA)}.\hskip 1em plus 0.5em minus 0.4em\relax IEEE, 2018, pp.
  1421--1428.

\bibitem{weng2022mtp}
X.~Weng, B.~Ivanovic, and M.~Pavone, ``{MTP: Multi-hypothesis Tracking and
  Prediction for Reduced Error Propagation},'' in \emph{Proceedings of the
  {IEEE} Intelligent Vehicles Symposium (IV)}, 2022, pp. 1218--1225.

\bibitem{bernreiter2019multiple}
L.~Bernreiter, A.~Gawel, H.~Sommer, J.~Nieto, R.~Siegwart, and C.~C. Lerma,
  ``{Multiple Hypothesis Semantic Mapping for Robust Data Association},''
  \emph{IEEE Robotics and Automation Letters (RA-L)}, vol.~4, no.~4, pp.
  3255--3262, 2019.

\bibitem{tan2022shape}
K.~Tan, Q.~Ji, L.~Feng, and M.~T{\"o}rngren, ``{Shape Estimation of a 3D
  Printed Soft Sensor Using Multi-Hypothesis Extended Kalman Filter},''
  \emph{IEEE Robotics and Automation Letters (RA-L)}, vol.~7, no.~3, pp.
  8383--8390, 2022.

\bibitem{heidenreichLaneSLAMSimultaneousPose2015}
T.~Heidenreich, J.~Spehr, and C.~Stiller, ``{LaneSLAM -- Simultaneous Pose and
  Lane Estimation Using Maps with Lane-Level Accuracy},'' in \emph{IEEE
  International Conference on Intelligent Transportation Systems (ITSC)}, 2015,
  pp. 2512--2517.

\bibitem{petrichAssessingMapbasedManeuver2014}
D.~Petrich, T.~Dang, G.~Breuel, and C.~Stiller, ``{Assessing Map-Based Maneuver
  Hypotheses Using Probabilistic Methods and Evidence Theory},'' in \emph{IEEE
  International Conference on Intelligent Transportation Systems (ITSC)}, 2014,
  pp. 995--1002.

\bibitem{krishanthPredictionRetrodictionAlgorithms2014}
K.~Krishanth, R.~Tharmarasa, T.~Kirubarajan, P.~Valin, and E.~Meger,
  ``{Prediction and Retrodiction Algorithms for Path-Constrained Targets},''
  \emph{IEEE Transactions on Aerospace and Electronic Systems}, 2014.

\bibitem{wijesomaObservabilityPathConstrained2006}
S.~Wijesoma, K.~W. Lee, and J.~I. Guzman, ``{On the Observability of Path
  Constrained Vehicle Localisation},'' in \emph{IEEE International Conference
  on Intelligent Transportation Systems (ITSC)}, 2006, pp. 1513--1518.

\bibitem{quddusCurrentMapmatchingAlgorithms2007}
M.~A. Quddus, W.~Y. Ochieng, and R.~B. Noland, ``{Current Map-Matching
  Algorithms for Transport Applications: State-of-the Art and Future Research
  Directions},'' \emph{Transportation Research Part C: Emerging Technologies},
  vol.~15, no.~5, pp. 312--328, 2007.

\bibitem{bernsteinIntroductionMapMatching}
D.~Bernstein and A.~Kornhauser, ``{An Introduction to Map Matching for Personal
  Navigation Assistants},'' US Transportation Collection.

\bibitem{leeConstrainedSLAMApproach2007}
K.~W. Lee, S.~Wijesoma, and J.~I. Guzmán, ``{A Constrained SLAM Approach to
  Robust and Accurate Localisation of Autonomous Ground Vehicles},''
  \emph{Robotics and Autonomous Systems}, vol.~55, no.~7, pp. 527--540, 2007.

\bibitem{liuGNSSTrackmapCooperative2014}
J.~Liu, B.-g. Cai, and J.~Wang, ``{A GNSS/Trackmap Cooperative Train
  Positioning Method for Satellite-Based Train Control},'' in \emph{IEEE
  International Conference on Intelligent Transportation Systems (ITSC)}, 2014,
  pp. 2718--2724.

\bibitem{jiangNewTrainIntegrity2020}
W.~Jiang, Y.~Liu, B.~Cai, C.~Rizos, J.~Wang, and Y.~Jiang, ``{A New Train
  Integrity Resolution Method Based on Online Carrier Phase Relative
  Positioning},'' \emph{IEEE Transactions on Vehicular Technology}, vol.~69,
  no.~10, pp. 10\,519--10\,530, 2020.

\bibitem{caoINSOdometerTrackmapaided2022}
Z.~Cao, J.~Liu, W.~Jiang, B.~Cai, and J.~Wang, ``{INS/Odometer/Trackmap-aided
  Railway Train Localization under GNSS Jamming Conditions},'' in
  \emph{Proceedings of the {IEEE} Intelligent Vehicles Symposium (IV)}, 2022,
  pp. 427--434.

\bibitem{chuGPSMEMSINS2013}
H.-J. Chu, G.-J. Tsai, K.-W. Chiang, and T.-T. Duong, ``{GPS/MEMS INS Data
  Fusion and Map Matching in Urban Areas},'' \emph{Sensors}, vol.~13, no.~9,
  pp. 11\,280--11\,288, 2013.

\bibitem{cossaboomAugmentedKalmanFilter2012}
M.~Cossaboom, J.~Georgy, T.~Karamat, and A.~Noureldin, ``{Augmented Kalman
  Filter and Map Matching for 3D RISS/GPS Integration for Land Vehicles},''
  \emph{International Journal of Navigation and Observation}, vol. 2012, pp.
  1--16, 2012.

\bibitem{brakatsoulas2005map}
S.~Brakatsoulas, D.~Pfoser, R.~Salas, and C.~Wenk, ``{On Map-Matching Vehicle
  Tracking Data},'' in \emph{Proceedings of the International Conference on
  Very Large Data Bases}, 2005, pp. 853--864.

\bibitem{yuanInteractiveVotingBasedMap2010}
J.~Yuan, Y.~Zheng, C.~Zhang, X.~Xie, and G.-Z. Sun, ``{An Interactive-Voting
  Based Map Matching Algorithm},'' in \emph{International Conference on Mobile
  Data Management}, 2010, pp. 43--52.

\bibitem{louMapmatchingLowsamplingrateGPS2009}
Y.~Lou, C.~Zhang, Y.~Zheng, X.~Xie, W.~Wang, and Y.~Huang, ``{Map-Matching for
  Low-Sampling-Rate GPS Trajectories},'' in \emph{Proceedings of the
  International Conference on Advances in Geographic Information Systems},
  2009, p. 352.

\bibitem{peng2020map}
C.~Peng and D.~Weikersdorfer, ``{Map as The Hidden Sensor: Fast Odometry-Based
  Global Localization},'' in \emph{IEEE International Conference on Robotics
  and Automation (ICRA)}, 2020, pp. 2317--2323.

\bibitem{kirubarajanTrackingGroundTargets1998}
T.~Kirubarajan, Y.~Bar-Shalom, K.~Pattipati, I.~Kadar, B.~Abrams, and E.~Eadan,
  ``{Tracking Ground Targets with Road Constraints Using an IMM Estimator},''
  in \emph{IEEE Aerospace Conference Proceedings}, vol.~5, 1998, pp. 5--12.

\bibitem{heirichBayesianTrainLocalization2016}
O.~Heirich, ``{Bayesian Train Localization with Particle Filter, Loosely
  Coupled GNSS, IMU, and a Track Map},'' \emph{Journal of Sensors}, vol. 2016,
  pp. 1--15, 2016.

\bibitem{lauerTrainLocalizationAlgorithm2014}
M.~Lauer and D.~Stein, ``{A Train Localization Algorithm for Train Protection
  Systems of the Future},'' \emph{IEEE Transactions on Intelligent
  Transportation Systems}, vol.~16, no.~2, pp. 970--979, 2014.

\bibitem{hasbergSimultaneousLocalizationMapping2012}
C.~Hasberg, S.~Hensel, and C.~Stiller, ``{Simultaneous Localization and Mapping
  for Path-Constrained Motion},'' \emph{IEEE Transactions on Intelligent
  Transportation Systems}, vol.~13, no.~2, pp. 541--552, 2012.

\bibitem{williamsNextETCSLevel2016}
C.~Williams, ``{The Next ETCS Level?}'' in \emph{IEEE International Conference
  on Intelligent Rail Transportation Proceedings (ICIRT)}, 2016, pp. 75--79.

\bibitem{kostrominovRFIDBasedNavigationSubway2020}
A.~M. Kostrominov, O.~N. Tyulyandin, A.~B. Nikitin, M.~N. Vasilenko, and A.~T.
  Osminin, ``{RFID-Based Navigation of Subway Trains},'' in \emph{IEEE
  East-West Design \& Test Symposiu}, 2020, pp. 1--6.

\bibitem{zhengTrainIntegratedPositioning2016}
W.~Zheng, S.~Ma, Z.~Hua, H.~Jia, and Z.~Zhao, ``{Train Integrated Positioning
  Method Based on GPS/INS/RFID},'' in \emph{Chinese Control Conference (CCC)},
  2016, pp. 5858--5862.

\bibitem{henselProbabilisticLandmarkBased2010}
S.~Hensel and C.~Hasberg, ``{Probabilistic Landmark Based Localization of Rail
  Vehicles in Topological Maps},'' in \emph{IEEE/RSJ International Conference
  on Intelligent Robots and Systems (IROS)}, 2010, pp. 4824--4829.

\bibitem{heirichStudyTrainSidePassive2017}
O.~Heirich, B.~Siebler, and E.~Hedberg, ``{Study of Train-Side Passive Magnetic
  Measurements with Applications to Train Localization},'' \emph{{Journal of
  Sensors}}, vol. 2017, pp. 1--10, 2017.

\bibitem{heirich2013velocity}
O.~Heirich, A.~Steingass, A.~Lehner, and T.~Strang, ``{Velocity and Location
  Information from Onboard Vibration Measurements of Rail Vehicles},'' in
  \emph{Proceedings of the International Conference on Information Fusion},
  2013, pp. 1835--1840.

\bibitem{wohlfeilVisionBasedRail2011}
J.~Wohlfeil, ``{Vision Based Rail Track and Switch Recognition for
  Self-Localization of Trains in a Rail Network},'' in \emph{Proceedings of the
  {IEEE} Intelligent Vehicles Symposium (IV)}, 2011, pp. 1025--1030.

\bibitem{naiTrainPositioningAlgorithm2017}
W.~Nai, Y.~Chen, X.~Zhang, X.~Lei, and D.~Dong, ``{A Train Positioning
  Algorithm Based on Inflexion Analysis by Using Sets of Distance Measurement
  Data Collected from On-Board Laser Ranging Equipments},'' in \emph{IEEE
  International Conference on Computer and Communications (ICCC)}, 2017, pp.
  901--904.

\bibitem{daoustLightEndTunnel2016}
T.~Daoust, F.~Pomerleau, and T.~D. Barfoot, ``{Light at the End of the Tunnel:
  High-Speed LiDAR-Based Train Localization in Challenging Underground
  Environments},'' in \emph{Conference on Computer and Robot Vision (CRV)},
  2016, pp. 93--100.

\bibitem{allottaLocalizationAlgorithmRailway2015}
B.~Allotta, P.~D'Adamio, M.~Malvezzi, L.~Pugi, A.~Ridolfi, and G.~Vettori, ``{A
  Localization Algorithm for Railway Vehicles},'' in \emph{IEEE International
  Instrumentation and Measurement Technology Conference (I2MTC)}, 2015, pp.
  681--686.

\bibitem{tschoppExperimentalComparisonVisualAided2019}
F.~Tschopp, T.~Schneider, A.~W. Palmer, N.~Nourani-Vatani, C.~Cadena,
  R.~Siegwart, and J.~Nieto, ``{Experimental Comparison of Visual-Aided
  Odometry Methods for Rail Vehicles},'' \emph{IEEE Robotics and Automation
  Letters (RA-L)}, vol.~4, no.~2, pp. 1815--1822, 2019.

\bibitem{tschoppVisualPositioningRailway2021a}
F.~Tschopp, ``{Visual Positioning for Railway Vehicles},'' Ph.D. dissertation,
  ETH Zurich, 2021.

\bibitem{winterGeneratingCompactGeometric2019a}
H.~Winter, S.~Luthardt, V.~Willert, and J.~Adamy, ``{Generating Compact
  Geometric Track-Maps for Train Positioning Applications},'' in
  \emph{Proceedings of the {IEEE} Intelligent Vehicles Symposium (IV)}, 2019,
  pp. 1027--1032.

\bibitem{acceptancegate2014}
Z.~Berman, ``Outliers rejection in kalman filtering — some new
  observations,'' in \emph{2014 IEEE/ION Position, Location and Navigation
  Symposium - PLANS 2014}, 2014, pp. 1008--1013.

\bibitem{visensor2014}
J.~Nikolic, J.~Rehder, M.~Burri, P.~Gohl, S.~Leutenegger, P.~T. Furgale, and
  R.~Siegwart, ``A synchronized visual-inertial sensor system with fpga
  pre-processing for accurate real-time slam,'' in \emph{2014 IEEE
  International Conference on Robotics and Automation (ICRA)}, 2014, pp.
  431--437.

\bibitem{furgale2013unified}
P.~Furgale, J.~Rehder, and R.~Siegwart, ``{Unified Temporal and Spatial
  Calibration for Multi-Sensor Systems.}'' in \emph{IEEE/RSJ International
  Conference on Intelligent Robots and Systems (IROS)}, 2013, pp. 1280--1286.

\bibitem{tschoppHoughMapIterative2021}
F.~Tschopp, C.~von Einem, A.~Cramariuc, D.~Hug, A.~W. Palmer, R.~Siegwart,
  M.~Chli, and J.~Nieto, ``{Hough$^2$Map – Iterative Event-Based Hough
  Transform for High-Speed Railway Mapping},'' \emph{IEEE Robotics and
  Automation Letters (RA-L)}, vol.~6, no.~2, pp. 2745--2752, 2021.

\end{thebibliography}
